\documentclass[10pt, a4paper]{article}

\usepackage[final]{lrec2026} 
\usepackage{algorithm}
\usepackage{algorithmic}
\usepackage{latexsym}
\usepackage{booktabs}
\usepackage[table]{xcolor}
\usepackage{subcaption}
\usepackage{pifont}
\newcommand{\redcross}{{\color{red}\ding{55}}}
\newcommand{\greencheck}{{\color{green}\ding{51}}}
\usepackage{xurl}
\usepackage{amsmath}
\usepackage{amssymb}
\usepackage{tikz}
\usetikzlibrary{positioning,trees}
\usepackage{subcaption}
\usepackage{pgfplots}
\usepackage{pgfplotstable}
\pgfplotsset{compat=1.18}

\title{Assessing LLM Reasoning Through Implicit Causal Chain Discovery in Climate Discourse}

\name{
\begin{tabular}{c}
Liesbeth Allein$^{\dagger*}$\thanks{$^{*}$ Current affiliation: Department of Electronics and Information Systems, Ghent University, Belgium. Contact: \texttt{liesbeth.allein@ugent.be}.}, Nataly Pineda-Castañeda$^{\ddagger}$, Andrea Rocci$^{\ddagger}$, \\
Marie-Francine Moens$^{\dagger}$
\end{tabular}
}

\address{
\begin{tabular}{c}
     $^{\dagger}$ Department of Computer Science, KU Leuven, Belgium\\
$^{\ddagger}$ Institute of Argumentation, Linguistics, and Semiotics (IALS), \\ Università della Svizzera italiana, Switzerland
\end{tabular}
}

\abstract{
How does a cause lead to an effect, and which intermediate causal steps explain their connection? This work scrutinizes the mechanistic causal reasoning capabilities of large language models (LLMs) to answer these questions through the task of implicit causal chain discovery. In a diagnostic evaluation framework, we instruct nine LLMs to generate all possible intermediate causal steps linking given cause-effect pairs in causal chain structures. These pairs are drawn from recent resources in argumentation studies featuring polarized discussion on climate change. Our analysis reveals that LLMs vary in the number and granularity of causal steps they produce. Although they are generally self-consistent and confident about the intermediate causal connections in the generated chains, their judgments are mainly driven by associative pattern matching rather than genuine causal reasoning. Nonetheless, human evaluations confirmed the logical coherence and integrity of the generated chains. Our baseline causal chain discovery approach, insights from our diagnostic evaluation, and benchmark dataset with causal chains lay a solid foundation for advancing future work in implicit, mechanistic causal reasoning in argumentation settings. 
 \\ \newline \Keywords{mechanistic causal reasoning, causal chain discovery, climate change, argumentation} }

\begin{document}

\maketitleabstract

\section{Introduction}

\textit{``The human cognitive system is built to see causation as governing how events unfold." \citep{sloman2015causality}} \\

Causal reasoning often hinges not on surface events, but on the underlying causal mechanisms that connect them. People are sensitive to these mechanisms when understanding, predicting, and evaluating real-world events and actions \citep{ahn1995role,ahn1996mechanism,walsh2011meaning}. They also perceive causal relations as stronger when structured in chains
\citep{keshmirian2024chain}. When faced with a causal relation like \textit{climate change $\rightarrow$ flooding} (i.e., \textit{climate change} causes \textit{flooding}), they tend to mentally break it down into its intermediate causal steps, e.g., \textit{climate change $\rightarrow$ excessive rainfall} and \textit{excessive rainfall $\rightarrow$ flooding}. This process eventually produces a causal chain-like structure, \textit{climate change $\rightarrow$ excessive rainfall $\rightarrow$ flooding} that reflects one possible pathway linking cause and effect.

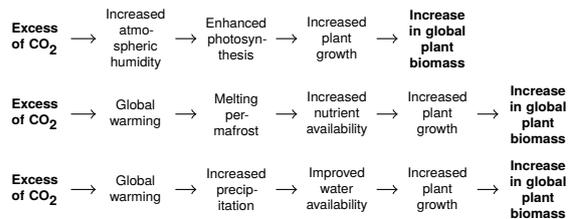
\begin{figure}[t]
    \centering
    \begin{subfigure}[t]{\columnwidth}
        \centering
        \caption{\textbf{Cause-effect relation} (\textit{cause} $\rightarrow$ \textit{effect}) in the statement \textit{``Additional CO\textsubscript{2} in the air has promoted growth in global plant biomass.''}:}
        \begin{tikzpicture}[
          grow=right,
          sibling distance=0.3cm,
          level distance=1.5cm,
          every node/.style={anchor=west,text width=2cm,font=\tiny,align=center},
          parent/.style={text width=2cm, font=\tiny,align=center},
          edge from parent path={
            [->] (\tikzparentnode.east) -- ++(0,0) -- (\tikzchildnode.west)
          }
        ]

        \node[parent] (root) {\textbf{Excess of CO\textsubscript{2}}}
          child {node (n1) {\textbf{Increase in global plant biomass}}};

        \draw[dashed, ->] (root.east) to (n1.west);

        \end{tikzpicture}
    \end{subfigure}
    \hfill
    \begin{subfigure}[t]{\columnwidth}
        \centering
        \caption{Three generated \textbf{forward mental simulations of the causal mechanism} linking the initial cause to the final effect in causal chain-like structures:}
        \begin{tikzpicture}[
          grow=right,
          sibling distance=0.3cm,
          level distance=0.8cm,
          every node/.style={anchor=west,text width=0.8cm,font=\tiny,align=center},
          parent/.style={text width=0.7cm, font=\tiny,align=center},
          edge from parent path={
            [->] (\tikzparentnode.east) -- ++(0,0) -- (\tikzchildnode.west)
          }
        ]

        \node[parent] (root1) {\textbf{Excess of CO\textsubscript{2}}}
          child {node (n1a) {Increased atmospheric humidity}
            child {node (n2a) {Enhanced photosynthesis}
              child {node (n3a) {Increased plant growth}
                  child {node (n4a) {\textbf{Increase in global plant biomass}}}}}
          };

        \begin{scope}[yshift=-1cm]
          \node[parent] (root2) {\textbf{Excess of CO\textsubscript{2}}}
            child {node (n1b) {Global warming}
              child {node (n2b) {Melting permafrost}
                child {node (n3b) {Increased nutrient availability}
                    child {node (n4b) {Increased plant growth}
                      child {node (n5b) {\textbf{Increase in global plant biomass}}}}}}
            };
        \end{scope}

        \begin{scope}[yshift=-2cm]
          \node[parent] (root3) {\textbf{Excess of CO\textsubscript{2}}}
            child {node (n1c) {Global warming}
              child {node (n2c) {Increased precipitation}
                child {node (n3c) {Improved water availability}
                    child {node (n4c) {Increased plant growth}
                      child {node (n5c) {\textbf{Increase in global plant biomass}}}}}}
            };
        \end{scope}

        \end{tikzpicture}
    \end{subfigure}
    \caption{This paper proposes the implicit causal chain discovery task for revealing multiple intermediate causal mechanisms (b) underlying a cause-effect relation that is put forward in an argumentative setting (a). It analyzes the chains and their intermediate causal relations state-of-the-art language models reconstruct and that explain the different latent causal pathways existing between an initial cause and final effect.}
    \label{fig:introduction_CC}
\end{figure}

In everyday discourse, causal relations are more often left implicit than other discourse relations \citep{Carbonell2005-ContrastMarks, Nadal2019-CausalitySpanish, Murray1997-Continuity, Zunino2014-CognitiveCausality, Zunino2011-Conectivas}. Speakers omit or simplify the mechanisms underlying causal relations, providing just enough information for listeners to infer meaning \citep{Grice1975-GRILAC-6}. They rely on shared knowledge or the assumption that the causal mechanisms are self-evident. This implicitness, however, leads to diverse interpretations among listeners \citep{allein2025interpretation}. It is therefore likely that a causal relation triggers different mental simulations of the underlying causal mechanistic pathways, with varying degrees of overlap in the intermediate events (Figure \ref{fig:introduction_CC}). 

And let it just be that \textbf{causality lies at the very center of argumentation in heavily-polarized topics} like climate change \citep{nicholson2014climate,cook2016rational,minnerop2019climate}. Divided skeptic and believer groups attribute and accept causality in climate events differently and maintain different causal mechanistic views that support their in-group views and refute those of the out-group \citep{bostrom2012causal,cole2025social}. While causality is a key driver in belief polarization regarding this topic \citep{cook2016rational,bayes2021motivated}, the implicitness of causality in discourse, together with the complexity of climate change systems (e.g., interdependent variables and feedback loops), greatly complicates identifying the points of division in causal thinking deeply embedded in different belief groups.

Within this context, \textbf{we consider implicit causal chain discovery a key task in argumentation}. This task involves inferring causal chain-like structures that explain the underlying causal mechanisms linking two events presented as a cause-effect (CE) pair during argumentation. In that respect, causal chains in this work reflect the \textit{forward mental simulations of causal mechanisms} that are potentially triggered in people's minds when reasoning over a given causal relation. This cognitive framing has practical implications for argumentation studies. The ability to infer detailed causal chains is highly valuable: making unexpressed causal links explicit can reveal how speakers rely on common ground, conversational implicatures, and implicitness in discourse. It may also assist in identifying weaknesses or gaps in causal explanations (such as missing links or unsupported assumptions), and developing stronger explanations by filling the gaps with additional evidence and details. Ultimately, the ability to evaluate and refine causal explanations makes causal chain discovery crucial for argument assessment, fallacy detection, production of counterarguments, and guidance in debate and critical thinking. 

\textbf{This paper scrutinizes the inherent capabilities of general-purpose and reasoning large language models (LLMs) to reconstruct implicit causal chains between events}. Like other causal discovery and reasoning tasks \citep{jin2023cladder,joshi-etal-2024-llms,jincan}, implicit causal chain discovery poses a highly complex challenge for LLMs. The frequent absence of explicit causal mechanisms is a reporting bias that affects the coverage and quality of the mechanistic causal knowledge represented in the text data used to train LLMs and build causal knowledge graphs \citep{10.1145/2509558.2509563,shwartz-choi-2020-neural,gao-etal-2023-chatgpt}. In addition, training data on polarized topics like climate change is rich in invalid or conflicting causality. The existence of multiple causal pathways through which cause and effect can arguably be linked also adds complexity. For example, one chain may involve \textit{excess in rainfall}, while another may include \textit{drought} (i.e., a deficiency in rainfall) as an intermediate step. Finally, determining the appropriate granularity of causal relations, the level of abstraction (i.e., molecular or observational), and the optimal chain length is far from straightforward and highly dependent on the specific causal relation under consideration.

\paragraph{Contributions} This paper illuminates the complexity of causal chain discovery and lays a solid foundation for advancing future work in causal chain discovery in argumentation.
The zero-shot causal chain generation approach we develop acts as a strong baseline, while our diagnostic evaluation with automatic and human assessment singles out valuable insights in LLM causal reasoning behavior. We also release a publicly available dataset of structurally consistent causal chains generated by multiple LLMs, offering a valuable resource for comparison and benchmarking\footnote{The code for reproducing the causal chain discovery experiments, the dataset of generated causal chains and intermediate CE pairs, and the human evaluation setup are available in a public Github repository: \url{https://github.com/laallein/implicit-causal-chain-discovery}.}.


\section{Causal Chain Discovery}\label{sec:experiments}

\subsection{Task Description} Given a \textit{CE pair} \( E_c \rightarrow E_f \), where \( E_c \) and \( E_f \) are noun phrases describing an event or action, represented as sequences of natural language tokens, and the arrow \( \rightarrow \) denotes the \textit{directionality} of causality from cause to effect, the objective is to generate a set of \( N \) \textit{causal chains} that connect \( E_c \) to \( E_f \) through a sequence of intermediate events. Note that \( E_c \) and \( E_f \) are used primarily as anchors, marking the start and end points for the causal chains.

Let \( \phi \) be a function parameterized by an LLM and $\lambda$ the template for formulating the input prompt, such that:
\[
\phi(\lambda(E_c, E_f)) \rightarrow \{C^{(1)}, C^{(2)}, \dots, C^{(N)}\}
\]

where each causal chain \( C^{(n)} \) is a sequence of the form:
\[
E_c \rightarrow E_{1}^{(n)} \rightarrow E_{2}^{(n)} \rightarrow \dots \rightarrow E_{M}^{(n)} \rightarrow E_f
\]

We let the LLM infer the appropriate value of $N$ for the given CE pair. Since each chain should be composed of at least three events, in accordance with the definition of a causal chain \citep{pearl2009causality}, $M>0$. The value of $M$ may vary across chains linking $E_c$ to $E_f$. Each intermediate event \( E_{m}^{(n)} \) represents a distinct step in the causal progression from \( E_c \) to \( E_f \). For the sake of brevity, we continue to refer to any event in a chain as $E_t$, with $t$ its position in the chain and $T$ the total length of the chain. Finally, each link in the chain, e.g., \( E_{t} \rightarrow E_{t+1}\), is treated as an \textit{intermediate CE pair} that can be independently analyzed and validated.

Note that we do not treat the Markov property \citep{pearl2009causality}, i.e., the conditional independence of non-adjacent events, as a necessary constraint when inferring causal chains in an argumentation context. 
In practice, human causal reasoning often violates the Markov property, as people tend to consider the broader chain of events rather than restrict themselves to strictly local dependencies \citep{rottman2016people}.

\subsection{Data}

We rely on annotated CE pairs from PolarIs3CAUS \citep{pineda2025polaris3} (95 pairs) and PolarIs4CAUS \citep{pineda2025polaris4} (181 pairs), which were manually extracted from English discussions on climate change in the PolarIS-3 (Reddit) and PolarIS-4 (X) datasets \cite{gajewska2024ethos}. 

Despite their modest number of CE pairs, PolarIs3CAUS and PolarIs4CAUS are particularly well-suited for causal chain discovery in this work and preferred over other existing resources \citep{mihuailua2013biocause,mirza-etal-2014-annotating,mostafazadeh-etal-2016-caters,dunietz-etal-2017-corpus,tan2022causal,romanou-etal-2023-crab,tan-etal-2023-recess}.

First, the CE annotations are high-quality, structurally consistent, and easy to understand (Table \ref{tab:data_sample}). Argumentation experts manually annotated the CE pairs as part of an argumentation-focused study on factual belief polarization \citep{pineda2026mapping}. The pairs were standardized through grammatical transformations and disambiguation procedures to ensure clarity outside their original context.

Second, the CE pairs appear in realistic argumentative discourse, where causal explanations are typically challenged or advanced as part of broader arguments. While they may not constitute full arguments, they may have argumentative or explanatory potential in broader disputes or dialogues. Moreover, this data offers insights into how causal reasoning operates in adversarial or deliberative contexts as it was extracted from two polarized belief groups, i.e., climate change believers and skeptics. 

Last, we expect that PolarIs3CAUS and PolarIs4CAUS are not included in the training data of the assessed LLMs as they were published after the data cut-off date of most LLMs.

\begin{table}[]
    \centering
    \setlength\extrarowheight{-2mm}
    \scriptsize
    \begin{tabular}{|p{2.6cm}p{1.45cm}p{0.1cm}p{1.85cm}|}
        \hline
        &&& \\
        \textbf{Message} & \textbf{Cause} & &  \textbf{Effect}  \\
        &&& \\
        \hline
        &&& \\
        Ocean acidification (pH inverse) is accelerating a direct result of CO2 emissions. & CO2 emissions & $\rightarrow$ & Ocean acidification \\
        &&& \\
        \hline
        &&& \\
        ``Business-as-usual" is leading to \#ClimateChange, \#biodiversity loss and the degradation of \#nature. To reverse this trend, we need to build a global \#CircularEconomy. & Business \newline \newline Business \newline \newline Business & $\rightarrow$ \newline \newline $\rightarrow$ \newline \newline $\rightarrow$ &  Climate change \newline \newline Biodiversity loss \newline Degradation of nature \\
        &&& \\
        \hline
        &&& \\
        Deforestation caused so much nutrient loss in the soils, eucalyptus is the only thing that has been able to successfully grow there. & Deforestation \newline \newline Nutrient loss in the soils & $\rightarrow$ \newline \newline $\rightarrow$ & Nutrient loss in the soils \newline Monoculture\\
        \hline
    \end{tabular}
    \caption{These three examples from the PolarIs4CAUS dataset with annotations of direct causal relations showcase the argumentative context of the causal relations (i.e., \textit{message}) and the structurally consistent formulation of cause and effect as noun phrases.}
    \label{tab:data_sample}
\end{table}

\subsection{Language Models}

We use
a broad set of open-source and proprietary LLMs for parameterizing $\phi$. \textbf{General-purpose LLMs} 
generate direct answers without showing intermediate reasoning steps: GPT4o \citep{hurst2024gpt}, Llama 3 70b \citep{grattafiori2024llama}, Mistral Nemo \citep{mistralnemo}, Llama 3.1 Nemotron \citep{bercovich2025llama}, Phi 4-mini \citep{abouelenin2025phi}, and Mixtral \citep{jiang2024mixtral}. \textbf{Reasoning LLMs} 
decompose complex problems into smaller problems and solve them in a step-by-step manner: o1, o1-mini \citep{jaech2024openai}, and DeepSeek R1 \citep{guo2025deepseek}\footnote{Implementation details can be found in the Appendix.}.

\subsection{Methodology}\label{CC_generation}

We implement a zero-shot prompting approach and instruct the LLM to generate all possible causal chains in a single step. We deliberately avoid a few-shot approach to prevent introducing biases related to the number of chains (i.e., $N$), the number of intermediate causal events in a chain (i.e., $M$), or the level of detail in each event.
We do not use chain-of-though prompting where we explicitly encourage step-by-step thinking, although
reasoning LLMs are expected to do this as part of their internal generation system. We also do not provide additional context through retrieval-augmented generation (RAG) as our goal is to assess the ability of LLMs to effectively draw on the causal knowledge encoded in their parameters rather than to evaluate their retrieval ability and use of external resources such as 
knowledge graphs (e.g., \citet{heindorf2020causenet, 10.5555/3491440.3491942}).

\begin{table}[]
    \centering
    \setlength\extrarowheight{-5mm}
    \scriptsize
    \begin{tabular}{|p{1.8cm}p{5cm}|}
        \hline
        & \\
        \textbf{Task} & \textbf{Prompts} \\
        & \\
        \hline
        & \\
        \textbf{Causal Chain Discovery} [$\lambda$] & A causal chain is a sequence of events in which each event directly causes the next, forming a connected series of cause-and-effect relations. Unfolding a causal chain means identifying and linking individual events. A step of the chain presents only one noun phrase containing the event. Unfold all possible causal chains that connect \texttt{\{$E_c$\}} (initial cause) to \texttt{\{$E_f$\}} (final effect) and separate the steps of the chain with the token \textless step\textgreater, and the chains with the token \textless chain\textgreater. \\
        & \\
        \hline
        & \\
        \textbf{Self-Consistency and Confidence [$\lambda_{A1}$]} & Answer with `yes' or `no' only. Does \texttt{\{$E_t$\}} cause \texttt{\{$E_{t+1}$\}}? \\
        & \\
        \hline
        & \\
        \textbf{Directionality of Causality [$\lambda_{A2}$]} & Answer with `yes' or `no' only. Does \texttt{\{$E_{t+1}$\}} cause \texttt{\{$E_t$\}}? \\
        & \\
        \hline
        & \\
        \textbf{Position Heuristics} & \textbf{[$\lambda_{A1-P}$]} Answer with `yes' or `no' only. Is \texttt{\{$E_{t+1}$\}} caused by \texttt{\{$E_t$\}}? \\
         & \textbf{[$\lambda_{A2-P}$]} Answer with `yes' or `no' only. Is \texttt{\{$E_t$\}} caused by \texttt{\{$E_{t+1}$\}}? \\
        & \\
        \hline
    \end{tabular}
    \caption{Overview of the template prompts $\lambda$ used in the causal chain discovery task and the evaluation setups. Curled brackets, \{\}, present input slots, which can be filled with a CE pair from the dataset, i.e., $E_c$ and $E_f$, or two consecutive events in a generated chain, i.e., $E_{t}$ and $E_{t+1}$.}
    \label{tab:prompts}
\end{table}

\begin{table*}[ht!]
    \centering
    \setlength{\tabcolsep}{0.9pt}
    \setlength\extrarowheight{0.5mm}
    \scriptsize
    \begin{minipage}{0.49\textwidth}
        \centering
        \begin{tabular}{llllllllll}
        & \parbox[t]{2mm}{\rotatebox[origin=l]{90}{\textbf{O1}}} & \parbox[t]{2mm}{\rotatebox[origin=l]{90}{\textbf{O1-mini}}} & \parbox[t]{2mm}{\rotatebox[origin=l]{90}{\textbf{GPT4o}}} & \parbox[t]{2mm}{\rotatebox[origin=l]{90}{\textbf{DeepSeek R1}}} & \parbox[t]{2mm}{\rotatebox[origin=l]{90}{\textbf{Llama 3.1 Nemo}}} & \parbox[t]{2mm}{\rotatebox[origin=l]{90}{\textbf{Llama 3 70b}}} & \parbox[t]{2mm}{\rotatebox[origin=l]{90}{\textbf{Mistral Nemo}}} & \parbox[t]{2mm}{\rotatebox[origin=l]{90}{\textbf{Mixtral}}} & \parbox[t]{2mm}{\rotatebox[origin=l]{90}{\textbf{Phi 4-mini}}} \\
        \toprule
        \rowcolor{lightgray} \multicolumn{10}{c}{\# Chains} \\
        \textbf{Total} & \textbf{865} & \textbf{478} & \textbf{485} & \textbf{581} & \textbf{716} & \textbf{500} & \textbf{363} & \textbf{575} & \textbf{346}\\
        \textbf{Per CE} &&&&&&&&& \\
        -- mean & 9.2 & 5.03 & 5.11 & 6.12 & 7.54 & 5.26 & 3.82 & 6.05 & 4.12\\
        -- std & 4.58 & 1.96 & 2.27 & 1.69 & 3.26 & 1.24 & 1.00 & 2.36 & 3.49\\
        -- min & 3 & 2 & 1 & 4 & 2 & 3 & 1 & 2 & 1\\
        -- max & 25 & 10 & 17 & 10 & 19 & 8 & 7 & 11 & 19\\
        \rowcolor{lightgray} \multicolumn{10}{c}{\# Intermediate CE Pairs} \\
        \textbf{Total} & \textbf{3,667} & \textbf{1,932} & \textbf{2,412} & \textbf{2,900} & \textbf{3,065} & \textbf{2,421} & \textbf{1,922} & \textbf{3,047} & \textbf{2,720} \\
        \textbf{Per chain} &&&&&&&&& \\
        --mean & 4.24 & 3.99 & 4.84 & 4.99 & 4.28 & 4.84 & 5.29 & 5.29 & 7.86 \\
        --std & 1.21 & 1.6 & 1.73 & 1.52 & 1.23 & 1.44 & 1.84 & 1.83 & 20.27 \\
        --min & 2 & 2 & 2 & 2 & 2 & 2 & 2 & 2 & 2 \\
        --max & 11 & 24 & 21 & 15 & 9 & 11 & 14 & 12 & 366 \\
        \rowcolor{lightgray} \multicolumn{10}{c}{Correlation \# chains and \# pairs per chain (= chain length)} \\
        Pearson's $r$ & -.19 & -.18 & -.16 & -.33 & -.21 & -.34 & -.39 & -.28 & /\\
        \bottomrule
        \end{tabular}
        \caption*{(a) PolarIs3CAUS}
    \end{minipage}
    \hfill
    \begin{minipage}{0.49\textwidth}
        \centering
        \begin{tabular}{llllllllll}
        & \parbox[t]{2mm}{\rotatebox[origin=l]{90}{\textbf{O1}}} & \parbox[t]{2mm}{\rotatebox[origin=l]{90}{\textbf{O1-mini}}} & \parbox[t]{2mm}{\rotatebox[origin=l]{90}{\textbf{GPT4o}}} & \parbox[t]{2mm}{\rotatebox[origin=l]{90}{\textbf{DeepSeek R1}}} & \parbox[t]{2mm}{\rotatebox[origin=l]{90}{\textbf{Llama 3.1 Nemo}}} & \parbox[t]{2mm}{\rotatebox[origin=l]{90}{\textbf{Llama 3 70b}}} & \parbox[t]{2mm}{\rotatebox[origin=l]{90}{\textbf{Mistral Nemo}}} & \parbox[t]{2mm}{\rotatebox[origin=l]{90}{\textbf{Mixtral}}} & \parbox[t]{2mm}{\rotatebox[origin=l]{90}{\textbf{Phi 4-mini}}} \\
        \toprule
        \rowcolor{lightgray} \multicolumn{10}{c}{\# Chains} \\
        \textbf{Total}& \textbf{1,689} & \textbf{932} & \textbf{874} & \textbf{1,163} & \textbf{3,714} & \textbf{943} & \textbf{744} & \textbf{1,216} & \textbf{911}\\
        \textbf{Per CE} &&&&&&&&& \\
        -- mean & 9.33 & 5.15 & 4.86 & 6.43 & 20.52 & 5.21 & 4.13 & 6.72 & 5.03 \\
        -- std & 4.67  & 1.5 & 1.59 & 1.97 & 34.12 & .95 & .94 & 2.89 & 6.86\\
        -- min & 2 & 1 & 1 & 3 & 3 & 4 & 2 & 3 & 1 \\
        -- max & 29 & 10 & 10 & 11 & 173 & 8 & 7 & 28 & 68  \\
        \rowcolor{lightgray} \multicolumn{10}{c}{\# Intermediate CE Pairs} \\
        \textbf{Total}& \textbf{7,308} & \textbf{3,610} & \textbf{4,026} & \textbf{5,606} & \textbf{17,926} & \textbf{4,336} & \textbf{4,015} & \textbf{6,343} & \textbf{10,099} \\
        \textbf{Per chain} &&&&&&&&& \\
        --mean & 4.33 & 3.81 & 4.58 & 4.82 & 4.82 & 4.60 & 5.40 & 5.22 & 11.06\\
        --std & 1.01 & 1.57 & 1.20 & 1.20 & 1.56 & 1.06 & 2.06 & 1.53 & 47.51  \\
        --min & 2 & 2 & 2 & 2 & 2 & 2 & 2 & 2 & 2 \\
        --max & 10 & 7 & 12 & 10 & 16 & 9 & 22 & 12 & 523 \\
        \rowcolor{lightgray} \multicolumn{10}{c}{Correlation \# chains and \# pairs per chain (= chain length)} \\
        Pearson's $r$ & / & -.12 & / & -.42 & -.37 & -.11 & -.29 & / & / \\
        \bottomrule
        \end{tabular}
        \caption*{(b) PolarIs4CAUS}
    \end{minipage}
    \caption{
    Statistics on the generated chains for CEs from PolarIs3CAUS (a) and PolarIs4CAUS (b): the number of generated chains for all CEs (total) and per CE, the number of intermediate CE pairs across all chains (total) and per chain, and statistically significant correlations between the number of chains generated for a CE and the length of those chains (Pearson's $r$; $p<.01$).
    }
    \label{tab:CC_inference}
\end{table*}

The input prompt $\lambda$ contains a causal chain definition, a task description with two slots for inserting $E_c$ and $E_f$, and formatting instructions (Table \ref{tab:prompts}). The prompt is designed to be domain-agnostic. Although directed acyclic graphs (DAGs) are the default representation of causal structures \citep{pearl2009causality}, we refrain from instructing LLMs to produce DAG outputs. This is because symbolic reasoning capabilities differ between LLMs \citep{tang2023large}, which we do not evaluate here. Importantly, the structurally consistent formulation of causes and effects in PolarIs3CAUS and PolarIs4CAUS allows us to impose grammatical formatting constraints, which improves consistency in the generation of intermediate events within the causal chains.

The prompt was iteratively refined through intensive manual prompt engineering. We removed ambiguities and non-essential information, with particular attention to the prompt's transferability across LLMs. The formatting instructions required multiple rounds of revision as the majority of the LLMs failed to follow them consistently. Nonetheless, the output from most LLMs required additional post-processing to parse the generated causal chains (e.g., removing intro).

\section{General Results}

The results in Table \ref{tab:CC_inference} show that the LLMs exhibit distinct causal chain discovery behavior. One key difference lies in their \textit{productivity}: models vary in the number of causal chains they generate for a given CE pair (i.e., $N$). However, this variation does not align with model type as there is no consistent distinction between general-purpose LLMs and reasoning LLMs.
Chain length (i.e., $T$), which reflects the depth of reasoning or \textit{verbosity} of a model, also does not clearly differentiate general-purpose LLMs from reasoning LLMs.

Interestingly, there is a statistically significant negative correlation ($p < .01$) between the number of chains generated for a CE pair and their average length, with Pearson's $r$ values ranging from $-.11$ to $-.42$ across models. In other words, as more chains are generated for a given CE pair, their average length tends to decrease. This finding is somewhat counter-intuitive as we would expect that generating more chains would lead to greater elaboration and detail, resulting in longer chains rather than shorter ones.

\section{Diagnostic Evaluation} \label{sec:evaluation}

\begin{figure}[h!]
    \centering
    \includegraphics[width=\linewidth]{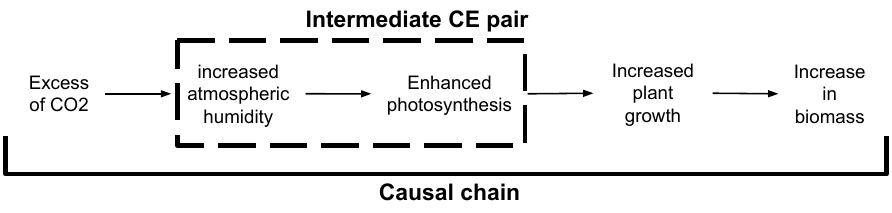}
    \caption{Example of a causal chain with four intermediate CE pairs.}
    \label{fig:example_chain}
\end{figure}

We continue to evaluate the generated causal chains, first \textit{at the level of the intermediate CE pairs in a chain} (i.e., $E_t \rightarrow E_{t+1}$), then \textit{at the level of the full chain} (i.e., $C$) (Figure \ref{fig:example_chain}). 

We design the evaluation setups such that we probe the inherent causal reasoning capabilities of LLMs in a structured, straightforward, and diagnostic manner. This way, we are able to identify specific points of success and failure in causal reasoning. 
Our setup substantially differs from previous work, which assessed causal reasoning capabilities by having LLMs answer questions about explicit causal relations in text \citep{chi2024unveiling}, compare causal relations \citep{ashwani2024cause,liu2024llms}, build and reason over causal graphs, usually in DAG structures, for causal relations explicitly mentioned in the input \citep{jin2023cladder,chen-etal-2024-clear,joshi-etal-2024-llms,wang-2024-causalbench}. 

Due to the high number of experiments and computing cost, we report the results from a single inference pass per model and evaluation configuration.

\subsection{Inside Causal Chains: Intermediate CE pairs}\label{evalCErels}  


We obtain the intermediate CE pairs by splitting all generated causal chains in their intermediate CE pairs. For instance, let $C = E_0 \rightarrow E_1 \rightarrow E_2 \rightarrow E_3$, then the three intermediate causal relations are $E_0 \rightarrow E_1$, $E_1 \rightarrow E_2$, and $E_2 \rightarrow E_3$. 
Detailed statistics on the intermediate CE pairs can be found in Table \ref{tab:CC_inference}. 

\subsubsection{Self-Consistency and Confidence in Causality (A1)}\label{assessment_1}

We evaluate the \textit{self-consistency} and \textit{confidence} of the LLMs regarding the intermediate CE pairs they generated during implicit causal chain discovery. We instruct the LLM to classify the relation of each intermediate CE pair as causal or non-causal by answering a yes/no question (\textit{`yes'} = causal, \textit{`no'} = non-causal) using the input prompt $\lambda_{A1}(E_t, E_{t+1})$ (Table \ref{tab:prompts}). A self-consistent and confident model should classify all intermediate CE pairs as causal.
\paragraph{Results} 
The majority of pairs are classified as causal, with minimal differences between general-purpose and reasoning LLMs. This suggests that the LLMs understood the causal chain discovery task well. They also demonstrate self-consistency in their reasoning and confidence regarding the causality between events. Furthermore, the results indicate that the quality and informativeness of the input prompt are sufficient, and that the prompt is transferable across different LLMs. Accordingly, we argue that our zero-shot prompting framework is suitable as a baseline for future causal chain discovery methods. 

\subsubsection{Directionality of Causality (A2)}\label{assessment_2}

A fundamental property of a causal chain is that its intermediate links exhibit a correct directional relationship, i.e., each event $E_t$ should be the cause of the subsequent event $E_{t+1}$; $E_t \rightarrow E_{t+1}$. To assess whether LLMs understand this property, we examine their understanding of \textit{temporal succession}, which is a key criterion of causation which asserts that causes should always precede their effects in time \citep{hume2000treatise,jonassen2008designing,grzymala2011time}. Consequently, a reversed relation, i.e., $E_{t+1} \rightarrow E_t$, violates this criterion and should be considered non-causal. 

Building on the experimental setup in A1, we present the LLMs with reversed CE pairs and instruct them to classify each as causal or non-causal using the input prompt $\lambda_{A2}(E_t, E_{t+1})$ (Table \ref{tab:prompts}). An LLM that has internalized the directionality of causal relations should consistently classify each reversed pair as non-causal.

\paragraph{Results} 
Around 50\% of the reversed CE pairs are still judged as causal, indicating that the models struggle to understand the directionality of causality. However, it is important to note that each CE pair is evaluated in isolation. The lack of context and minimal refinement of the events can result in pairs being considered causal in both the original and reversed order. This outcome is reasonable in some real-world settings. For example, \textit{harm to trees} $\rightarrow$ \textit{pollution} is valid and is explained through the chain \textit{harm to trees} $\rightarrow$ \textit{loss of forests} $\rightarrow$ \textit{soil erosion} $\rightarrow$  \textit{release of sediments and pollutants into water bodies} $\rightarrow$ \textit{water pollution}\footnote{\url{https://www.emission-index.com/deforestation/water-pollution}}. Conversely, \textit{pollution} $\rightarrow$ \textit{harm to trees} is also valid and is motivated by \textit{air pollution} $\rightarrow$ \textit{tree damage} $\rightarrow$  \textit{smaller trees and increased tree growth} $\rightarrow$ \textit{harm to trees}\footnote{\url{https://www.nps.gov/articles/000/how-to-assess-air-pollutions-impacts-on-forests.htm}}. The topic of climate change also contributes to the complexity as certain climate-related events can feed back into themselves, creating cyclic causality. For instance, the Ice-Albedo feedback is a cyclic process where melting ice reduces surface reflectivity (albedo), causing more solar energy to be absorbed, which leads to further warming and more ice melt. This stresses the importance of generating multiple causal chains for explaining causal mechanisms connecting events. 

\subsubsection{Fallacy in Causal Reasoning: Position Heuristics (A3)}\label{assessment_3}

LLMs have been shown to have internalized causal patterns during training (e.g., the event representing the cause is more frequently positioned before the event representing the effect in a text), which they reuse to answer causal questions quickly and efficiently without performing explicit causal inference \citep{jin2023cladder,zevceviccausal}.

We evaluate whether LLMs maintain consistent causal judgments when the surface form of prompts is 
altered but the semantic content is maintained by switching from active (\textit{``Does $E_t$ cause $E_{t+1}$?”}) to passive voice (\textit{``Is $E_{t+1}$ caused by $E_{t}$?”}). We instruct them to classify each $E_t \rightarrow E_{t+1}$ as causal or non-causal using the passive formulation of $\lambda_{A1}$, i.e., $\lambda_{A1-P}(E_t, E_{t+1})$. We repeat this, but now for the reversed pair $E_{t+1} \rightarrow E_t$ using the passive formulation of $\lambda_{A2}$, i.e., $\lambda_{A2-P}(E_t, E_{t+1})$ using input prompt. See Table \ref{tab:prompts} for $\lambda_{A1-P}$ and $\lambda_{A2-P}$. LLMs that do not resort to \textit{position heuristics} \citep{joshi-etal-2024-llms} and \textit{amortized causal reasoning} should consistently give the same answers to prompt pairs $\lambda_{A1}$ and $\lambda_{A1-P}$, and to $\lambda_{A2}$ and $\lambda_{A2-P}$. 

Consistency is measured using two metrics: Jaccard dissimilarity, which measures the dissimilarity between sets by normalizing the overlap of \textit{`yes'} responses and subtracting the result from one, and Hamming distance, which captures the proportion of differing answers across prompt pairs. Low scores on both metrics indicate strong alignment between active and passive prompts,
suggesting a 
robust understanding of the underlying causal relations.

\paragraph{Results}

For $\lambda_{A1}$ and $\lambda_{A1-P}$, disagreement between active and passive formulations is relatively low, with Jaccard dissimilarity and Hamming distance values ranging from 0.09 to 0.24 for both PolarIs3CAUS and PolarIs4CAUS. This suggests that the models exhibit a reasonable degree of consistency in their causal judgments, even when the linguistic framing is altered.
However, for $\lambda_{A2}$ and $\lambda_{A2-P}$, disagreement increases substantially, with values ranging from 0.25 to 0.51. This indicates a marked drop in consistency when evaluating reversed, incorrect causal relations. The elevated disagreement highlights the susceptibility of LLMs to position heuristics.

These results underscore a key limitation in current LLMs. Even those designed to perform well on reasoning tasks struggle to maintain stable causal inferences when superficial aspects of the prompt are modified. This prompt-sensitivity becomes especially problematic in contexts with strong polarization. LLMs would be susceptible to framing effects \citep{feldman2018climate}, inferring different causal chains depending on how a question is framed and potentially reinforcing existing ideological biases. Moreover, if an LLM cannot maintain consistent causal judgments across prompt variants, it may produce contradictory explanations for a single CE pair.

\begin{figure}[t]
    \centering
    \begin{subfigure}[b]{0.99\linewidth}
        \centering
        \includegraphics[width=\linewidth]{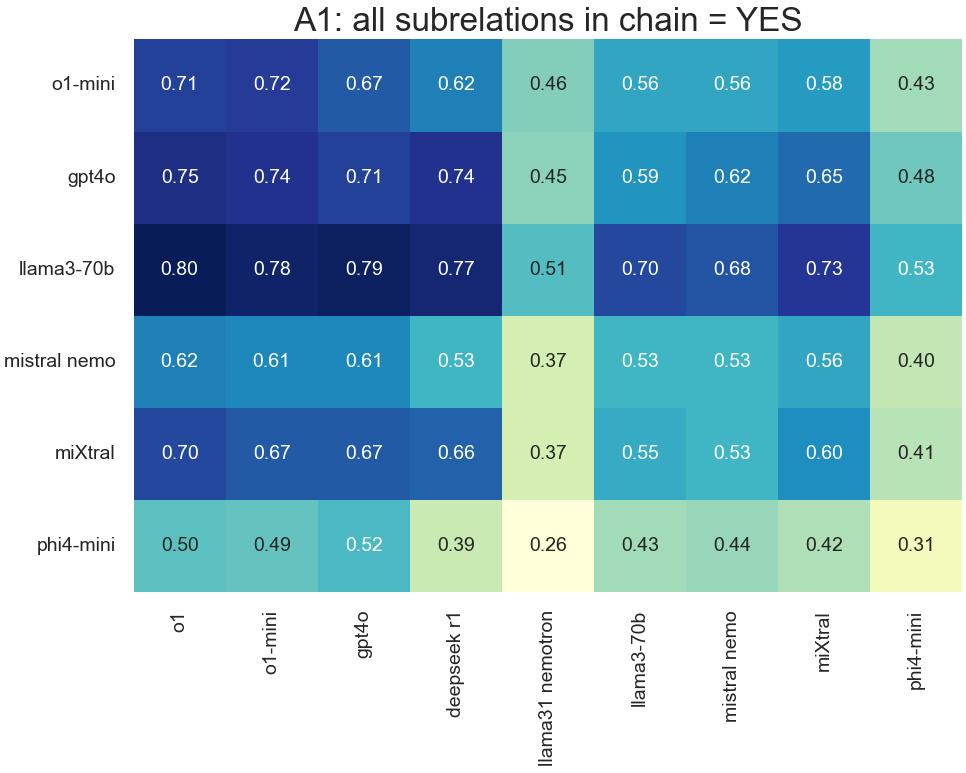}
        \label{fig:subfig1}
    \end{subfigure}
    \begin{subfigure}[b]{0.99\linewidth}
        \centering
        \includegraphics[width=\linewidth]{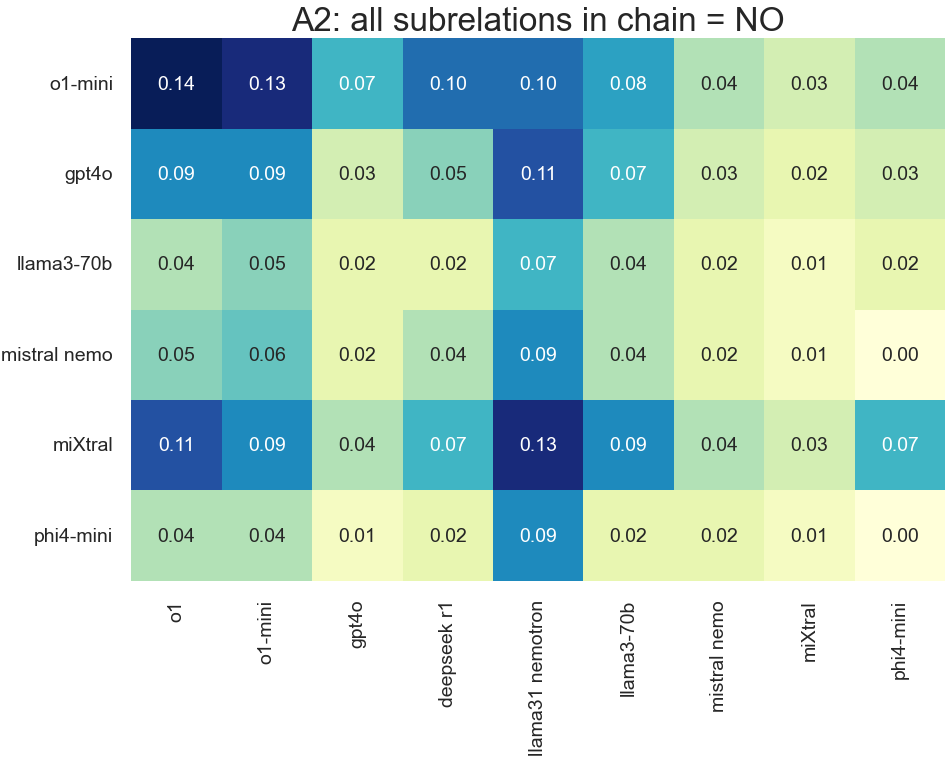}
        \label{fig:subfig2}
    \end{subfigure}
    \caption{Cross-model evaluation of the integrity of the chains generated for the CE pairs from PolarIs4CAUS. The results for the A1 setup are shown on top, those for the A2 setup below. In both figures, the LLMs that generated the chains are on the x-axis, the LLMs that evaluate those chains are on the y-axis, and the values report the proportion of chains that are considered valid by the LLM. }
    \label{fig:chain_validity}
\end{figure}

\begin{figure}[ht!]
\centering

\begin{subfigure}[t]{0.24\textwidth}
  \centering
  \begin{minipage}{\textwidth}
    \begin{tikzpicture}[xscale=1.7, yscale=0.2]
      \draw[-] (-0.6,-0.5) -- (0.6,-0.5);
      \draw[black, thin] (0,-0.5) -- (0,4);
      \foreach \y/\label in {3/o1, 2/Llama Nemo, 1/Mistral Nemo, 0/Mixtral} {
        \node[left, font=\tiny] at (-0.7,\y+0.25) {\scalebox{0.9}{\label}};
      }
      \draw[fill=blue] (-0.15,0.2) rectangle (-0.07,0.3);   
      \draw[fill=blue] (-.27,1.2) rectangle (-.13,1.3);     
      \draw[fill=blue] (.29,2.2) rectangle (.37,2.3);       
      \draw[fill=blue] (0.08,3.2) rectangle (.14,3.3);      
      \foreach \x in {-0.5,0,0.5} {
        \draw[thick] (\x,-0.2) -- (\x,-0.7) node[below, font=\tiny] {\x};
      }
      \node[above right, font=\small] at (-1, 3.2) {\scalebox{0.8}{\textbf{$\lambda_{A1}$}}};
      \node[left, font=\small] at (-0.75, -1.7) {\scalebox{0.8}{$r$}};
    \end{tikzpicture}


    \begin{tikzpicture}[xscale=1.7, yscale=0.2]
      \draw[-] (-0.6,-0.5) -- (0.6,-0.5);
      \draw[black, thin] (0,-0.5) -- (0,3);
      \foreach \y/\label in {2/Deepseek, 1/Llama Nemo, 0/Mixtral} {
        \node[left, font=\tiny] at (-0.7,\y+0.25) {\scalebox{0.9}{\label}};
      }
      \draw[fill=blue] (.09,0.2) rectangle (.13,0.3);   
      \draw[fill=blue] (-.29,1.2) rectangle (-.13,1.3);     
      \draw[fill=blue] (-.11,2.2) rectangle (-.08,2.3);       
      \foreach \x in {-0.5,0,0.5} {
        \draw[thick] (\x,-0.2) -- (\x,-0.7) node[below, font=\tiny] {\x};
      }
      \node[above right, font=\small] at (-1, 3.2) {\scalebox{0.8}{\textbf{$\lambda_{A2}$}}};
      \node[left, font=\small] at (-0.75, -1.7) {\scalebox{0.8}{$r$}};
    \end{tikzpicture}
  \end{minipage}
  \caption{Chain length and \% valid intermediate CE pairs.}
\end{subfigure}
\hfill
\begin{subfigure}[t]{0.23\textwidth}
  \centering
  \begin{minipage}{\textwidth}
    \begin{tikzpicture}[xscale=1.7, yscale=0.2]
      \draw[-] (-0.6,-0.5) -- (0.6,-0.5);
      \draw[black, thin] (0,-0.5) -- (0,1);
      \foreach \y/\label in {0/Llama Nemo} {
        \node[left, font=\tiny] at (-0.7,\y+0.25) {\scalebox{0.9}{\label}};
      }
      \draw[fill=blue] (-0.48,0.2) rectangle (-0.24,0.3);     
      \foreach \x in {-0.5,0,0.5} {
        \draw[thick] (\x,-0.2) -- (\x,-0.7) node[below, font=\tiny] {\x};
      }
      \node[above right, font=\small] at (-1, 3.2) {\scalebox{0.8}{\textbf{$\lambda_{A1}$}}};
        \node[left, font=\small] at (-0.75, -1.7) {\scalebox{0.8}{$r$}};
    \end{tikzpicture}


    \begin{tikzpicture}[xscale=1.7, yscale=0.2]
      \draw[-] (-0.6,-0.5) -- (0.6,-0.5);
      \draw[black, thin] (0,-0.5) -- (0,3);
      \foreach \y/\label in {2/Deepseek, 1/Llama 70b, 0/Llama Nemo} {
        \node[left, font=\tiny] at (-0.7,\y+0.25) {\scalebox{0.9}{\label}};
      }
      \draw[fill=blue] (.23,0.2) rectangle (.56,0.3);   
      \draw[fill=blue] (.24,1.2) rectangle (.35,1.3);     
      \draw[fill=blue] (.25,2.2) rectangle (.41,2.3);       
      \foreach \x in {-0.5,0,0.5} {
        \draw[thick] (\x,-0.2) -- (\x,-0.7) node[below, font=\tiny] {\x};
      }
      \node[above right, font=\small] at (-1, 3.2) {\scalebox{0.8}{\textbf{$\lambda_{A2}$}}};
      \node[left, font=\small] at (-0.75, -1.7) {\scalebox{0.8}{$r$}};
    \end{tikzpicture}
  \end{minipage}
  \caption{Number of chains and \% valid intermediate CE pairs.}
\end{subfigure}

\caption{Pearson correlation between chain length and the proportion (\%) of \textit{causal} intermediate CE pairs (a), and between the number of chains per CE pair and the proportion (\%) of \textit{causal} intermediate CE pairs (b). The y-axis includes models with significant correlation in their generations ($p < .01$). The x-axis shows the range of $r$ values. Causality of pairs is evaluated using $\lambda_{A1}$ and $\lambda_{A2}$.}
\label{fig:correlations}
\end{figure}
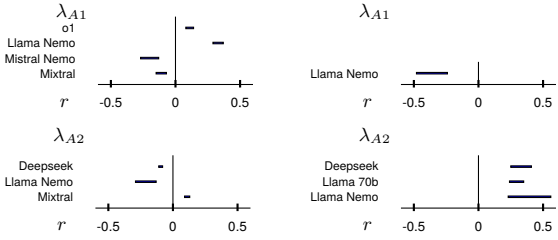

\subsection{On Causal Chains} \label{evalCC}

\subsubsection{Self- and Cross-Model Assessment of Causal Chain Integrity (A4) }\label{crossmodeleval}

We evaluate the \textit{integrity} of the generated causal chains. A chain's integrity is preserved only if all of its intermediate links are causal, i.e., answer to $\lambda_{A1}(E_t, E_{t+1})$ = \textit{`yes'} and $\lambda_{A2}(E_t, E_{t+1})$ = \textit{`no'}. It is violated if at least one link is non-causal. 
We let LLMs evaluate their own generations and of those others. 
\paragraph{Results} Figure \ref{fig:chain_validity} shows the proportion of causal chains generated by an LLM with preserved integrity as assessed by the model itself or another. Reasoning models
seem to produce a higher proportion of valid chains than general-purpose LLMs, except for GPT4o. 
This suggests that reasoning models are overall better at identifying clear causal relations.

We further examine the correlation between chain length and the proportion of intermediate CE pairs classified as causal through $\lambda_{A1}$ and $\lambda_{A2}$. \citet{xiong-etal-2022-reco} showed that performance on causal discovery tasks tends to decline with longer chains as they require more complex reasoning. We therefore hypothesize a negative correlation.
However, as shown in Figure \ref{fig:correlations}, correlation is not consistently negative across models. Chain length seems to affect the quality of causal discovery, though the direction and strength of this effect vary across models.

Extending this analysis, we investigate whether generating more chains for a given CE pair dilutes quality. For this, we assess the correlation between the number of chains generated for a CE pair and the proportion of intermediate CE pairs in those chains deemed causal by the LLMs (again using $\lambda_{A1}$ and $\lambda_{A2}$). The results in Figure \ref{fig:correlations} show that the impact of chain quantity on quality differs across model as some improve with more chains while others worsen.

Although assessing the integrity and quality of a chain by the validity of its intermediate CE pairs provides useful insights, we note that it does not fully account for certain complexities specific to causal chains, such as transitive inference, scene drift, and threshold effects \citep{xiong-etal-2022-reco}. For example, the following causal sequence highlights the subtlety of threshold effects: \textit{cold $\rightarrow$ vasoconstriction $\rightarrow$ increased blood pressure $\rightarrow$ stroke $\rightarrow$ death}. 

\begin{table*}[h]
    \centering
    \setlength\extrarowheight{-1.5mm}
    \scriptsize
    \begin{tabular}{p{11.2cm}p{1.25cm}p{0.7cm}p{1cm}}
    & \textbf{LLM Eval} & \multicolumn{2}{c}{\textbf{Human Eval}} \\
    \cmidrule{2-4}
    & Integrity & Integrity & Coherence \\
    \textbf{rich countries} $\rightarrow$ high waste generation $\rightarrow$ methane emissions from landfills $\rightarrow$ greenhouse gas accumulation $\rightarrow$ \textbf{climate change} & \greencheck \greencheck \greencheck \greencheck \redcross \redcross \redcross & \greencheck \greencheck \greencheck \greencheck & \greencheck \greencheck \greencheck \greencheck \\
    \textbf{rich countries} $\rightarrow$ industrialization $\rightarrow$ fossil fuel combustion $\rightarrow$ enhanced greenhouse effect $\rightarrow$ rich countries $\rightarrow$ overconsumption $\rightarrow$ resource depletion $\rightarrow$ environmental degradation $\rightarrow$ \textbf{climate change} & \greencheck \redcross \redcross \redcross \redcross \redcross \redcross \redcross & \greencheck \greencheck \redcross \redcross & \greencheck \greencheck \redcross \redcross\\
    \midrule
    \textbf{climate change} $\rightarrow$ earlier seasonal changes $\rightarrow$ mismatched reproductive cycles $\rightarrow$ population decline $\rightarrow$ \textbf{species at risk of extinction} & \greencheck \greencheck \greencheck \redcross \redcross \redcross \redcross & \greencheck \greencheck \greencheck \redcross & \greencheck \greencheck \greencheck \redcross \\
    \textbf{climate change} $\rightarrow$ ocean acidification $\rightarrow$ death of shell-forming organisms $\rightarrow$ disruption of marine food webs $\rightarrow$ collapse of fish populations $\rightarrow$ \textbf{species at risk of extinction} & \greencheck \redcross \redcross \redcross \redcross \redcross \redcross \redcross & \greencheck \greencheck \greencheck \greencheck & \greencheck \greencheck \greencheck \greencheck\\
    \end{tabular} 
    \caption{Examples of chain sets with agreement (top) and disagreement (bottom) between LLMs and argumentation experts.}
    \label{tab:humaneval}
\end{table*}

\subsubsection{Human Assessment of Causal Chain Validity and Logical Coherence (A5)} \label{humaneval}

The last evaluation setup includes a focused human evaluation with argumentation experts. We explore whether their judgments on the integrity of generated causal chains aligns with those of the LLMs. They also assess the logical coherence of the generated causal chains. We end by surveying them about the complexity of evaluating causal chains to gain insights into the feasibility of obtaining higher-quality chains from non-expert annotators, i.e., people who do not have extensive background on the topic of the causes and effect (in this study, climate change)\footnote{Details on chain selection procedure and screenshots of annotation forms are included in the Appendix.}.

\paragraph{Sample selection} The human evaluation focuses on a curated selection of 36 causal chains for 18 CE pairs from PolarIs4CAUS; two chains per CE. All chains have been generated by o1. We select the most agreed-upon \textit{maintained} and \textit{violated} chain per CE pair in terms of integrity, making sure that their lengths are approximately equal. Table \ref{tab:humaneval} shows two sets of chain pairs.

\paragraph{Evaluation setup} The human evaluation took place in a controlled, in-person setting with 10 master's and PhD students from a communication faculty. Each chain was independently judged by four participants. Participation was voluntary, and while students were informed the study concerned causal chain reasoning, they were not told the chains were LLM-generated to avoid bias. After receiving task instructions, which included a definition of a causal chain and two illustrative examples, they annotated six pairs of chains, labeled randomly as Chain A and Chain B. For each chain, they judged whether it maintained \textit{chain integrity} and was \textit{logically coherent}, defined as a plausible sequence where each step follows naturally from the previous one. At the end, participants reflected on their confidence and rated the difficulty of the annotation task on a 5-point Likert scale. They were also asked whether they believe they can construct a valid, coherent chain on climate change. If yes, they described how their chain would compare in length and detail to those evaluated; if no, they were asked to explain why.

\paragraph{Results} 
Based on the majority votes for integrity and logical coherence, the quality of the generated causal chains is rated relatively high: participants confirmed the integrity of 27 out of 36 chains and marked 24 as coherent. However, LLM judgments align poorly with human assessments as they disagreed on the integrity for half of the chains. Despite the cognitive demands of the task and low inter-annotator agreement (Fleiss' $\kappa = .084$ for integrity; $\kappa = .035$ for coherence), participants expressed confidence in their ability to construct valid and coherent causal chains. Nonetheless, clear instructions on the desired depth and detail are are essential for consistency as some participants anticipated producing shorter and simpler chains, while others expected theirs to be longer and more detailed.



\section{Related Work}

Unraveling complex causal structures beyond direct causal relations is central to advancing the modeling and analysis of real-world commonsense causality \citep{cui-etal-2024-odyssey}. Prior work has built causal graphs between CE pairs, for instance, using causal knowledge graphs to retrieve relevant events \citep{du-etal-2021-excar,xu2024exploring}. Specifically on causal chains, \citet{xiong-etal-2022-reco} proposed a framework to estimate the reliability of short causal chains, addressing transitive problems such threshold effect and scene drift. \citet{du-etal-2022-e} explained causal relations using single conceptual explanations. Closest to our work, \citet{ho2023wikiwhy} generated multi-hop causal chains to explain causal relations sourced from Wikipedia. However, their event representation is inconsistent, mixing full sentences and noun phrases, and they produce only one chain per relation, limiting coverage of alternative pathways. In contrast, this paper investigates the mechanistic causal reasoning capabilities of LLMs by evaluating their ability to generate multiple implicit causal chains that explain the causal relations between events in climate discourse.

Regarding the modeling of polarization in discourse, argumentation mining, including framing analysis and stance detection, has proven effective \citep{bianchi-etal-2021-sweat,sinno-etal-2022-political,hofmann-etal-2022-modeling,irani2024argusense}. Within argumentation, causality has been primarily studied in three main contexts: (i) discovery of explicit causal relations between events in text \citep{al2023new,tan-etal-2023-recess}, (ii) identification of causal relations between arguments \citep{jo2021classifying,saadat-yazdi-etal-2023-uncovering,bezou-vrakatseli-etal-2025-large}, and (iii) causal analysis of actions in persuasive conversations \citep{zeng2025causal}. 

\section{Conclusion}

Implicit causal chain discovery is key for studying common ground in causal thinking. By surfacing the underlying reasoning behind opposing views, such discovery may help individuals recognize how others arrive at their conclusions and identify areas of agreement. It also reveals overgeneralizations, false equivalences, and manipulative rhetoric rooted in faulty causal logic. This work sets a baseline for causal chain discovery that is transferable across LLMs, but also produces chains whose integrity and coherence are confirmed by argumentation experts through human evaluation. Our diagnostic evaluation of LLM behavior showed that LLMs are self-consistent and confident about the causality between events. However, they are vulnerable to minor changes in causal formulations, highlighting a reliance on position heuristics and a lack of robustness.

\paragraph{Future work} Important challenges remain. Although our evaluation design is domain-agnostic, we expect different causal discovery behavior and quality across domains. Some domains involve more complex or implicit causal mechanisms, reducing the coverage of relevant knowledge in the training data. In domains where causality is heavily contested, LLMs and causal knowledge graphs are more likely to encode conflicting or invalid causal associations, which further complicates the evaluation of causal validity. While RAG is a promising follow-up baseline, it may struggle to retrieve, align, and evaluate causal relations effectively due to subtle mismatches in the linguistic formulation of CE pairs.

\section*{Acknowledgements} 

This work has been funded by the Research Foundation - Flanders (FWO) under grant G0L0822N (Liesbeth Allein, Marie-Francine Moens) and the Swiss National Science Foundation (SNSF) under grant 209674 (Nataly Pineda-Castañeda, Andrea Rocci) through the CHIST-ERA project ``iTRUST Interventions against Polarisation in Society for Trustworthy Social Media. From Diagnosis to Therapy''.

\section*{Bibliographical References}\label{sec:reference}

\bibliographystyle{lrec2026-natbib}
\bibliography{lrec2026-example}


\end{document}